\documentclass{article}
\usepackage{tabularx}
\usepackage{PRIMEarxiv}
\usepackage{natbib}
\usepackage[utf8]{inputenc} % allow utf-8 input
\usepackage[T1]{fontenc}    % use 8-bit T1 fonts
\usepackage{hyperref}       % hyperlinks
\usepackage{url}            % simple URL typesetting
\usepackage{booktabs}       % professional-quality tables
\usepackage{amsfonts}       % blackboard math symbols
\usepackage{nicefrac}       % compact symbols for 1/2, etc.
\usepackage{microtype}      % microtypography
\usepackage{lipsum}
\usepackage{fancyhdr}       % header
\usepackage{graphicx}       % graphics
\graphicspath{{media/}}     % organize your images and other figures under media/ folder
\usepackage{longtable}
\usepackage{float}
\usepackage{enumitem}
\usepackage{amsmath}
\usepackage{graphicx}

\title{GLiNER multi-task: Generalist Lightweight Model for Various Information Extraction Tasks
}
\author{
  Ihor Stepanov \\
  Knowledgator Engineering LTD \\
  London \\
  \texttt{ingvarstep@knowledgator.com} \\
  \And
  Mykhailo Shtopko \\
  Knowledgator Engineering LTD \\
  London\\
  \texttt{mykhailoshtopko@gmail.com} \\
}
\begin{document}
\maketitle
\begin{abstract}
Information extraction tasks require both accurate, efficient, and generalisable models. Classical supervised deep learning approaches can achieve the required performance, but they need large datasets and are limited in their ability to adapt to different tasks. On the other hand, large language models (LLMs) demonstrate good generalization, meaning that they can adapt to many different tasks based on user requests. However, LLMs are computationally expensive and tend to fail to generate structured outputs. In this article, we will introduce a new kind of GLiNER model that can be used for various information extraction tasks while being a small encoder model. Our model achieved SoTA performance on zero-shot NER benchmarks and leading performance on question-answering, summarization and relation extraction tasks. Additionally, in this article, we will cover experimental results on self-learning approaches for named entity recognition using GLiNER models. \end{abstract}
% keywords can be removed

\section{Introduction}\label{sec:introduction}
Information extraction (IE) is an important discipline for many domains, including science \citep{Hong2021ChallengesAA}, business \citep{Skalicky2022BusinessDI}, public administration \citep{Siciliani2023OIE4PAOI} and much more. Considering the importance and influence of domains where information extraction approaches are applied, requirements for IE models are strict. They should be both efficient, meaning they require less computational resources to process more unstructured data per time. High accuracy is a prior requirement, especially for disciplines like biomedicine or business, where incorrect data can cause harm to people or lead to financial losses. Taking into account the diversity of information extraction tasks and the context in which they should be solved, models for IE should be effortlessly adaptable to new tasks and domains. 

The history of information extraction (IE) methods dates back to the early days of artificial intelligence and computational linguistics in the 1960s and 1970s \citep{Sager1980NaturalLI}. Early IE systems were rule-based, relying on manually crafted patterns and heuristics to extract specific types of information from text. These systems were often brittle and domain-specific, requiring extensive effort to update and maintain. The advent of machine learning in the 1990s brought significant advancements, enabling the development of more flexible and robust IE systems. Statistical methods, such as Hidden Markov Models (HMMs) \citep{seymore1999learning} and Conditional Random Fields (CRFs) \citep{mcdonald2005identifying}, allowed for more generalizable approaches to entity recognition and relation extraction.

In recent years, the rise of deep learning has revolutionized the field, with neural network-based models achieving state-of-the-art performance across various IE tasks. The introduction of transformer architectures, particularly BERT (Bidirectional Encoder Representations from Transformers), marked a breakthrough in language understanding. BERT's ability to understand the context in both directions (left-to-right and right-to-left) has dramatically improved the accuracy, especially for tasks like named entity recognition (NER), relation extraction, and other IE tasks \citep{devlin2018bert}. Following BERT, models like GPT-3 and newer versions (Generative Pre-trained Transformers) have pushed the boundaries even further. These models, with their massive scale and ability to generate coherent and contextually relevant text, have opened new possibilities for IE, enabling systems to handle more complex and nuanced information extraction tasks, adapting to user requirements \citep{brown2020language, openai2024gpt4}. These modern methods not only enhance accuracy and efficiency but also demonstrate greater adaptability to new domains and tasks, addressing many of the limitations of earlier techniques \citep{nadeau2007survey, parsing2009speech}.

However, generative approaches have their own limitations. First, they are not computationally efficient and need to generate output even if it is already presented in the text. Moreover, they tend to fail to provide structured output, which can lead to errors and decrease interpretability, which is very important for such domains as biomedicine. In this work, we will discuss our new approach based on GLiNER work \citep{zaratiana2023gliner}. It demonstrated SoTA results on zero-shot NER and other information extraction benchmarks while being more efficient and controllable.

\section{Methods}\label{seq:methods}

\subsection{Models architecture}
Our model is based on GLiNER token classification architecture. In comparison to the original work, it classifies tokens instead of spans, and it enables longer sequence extraction that is important for such tasks as long entity extraction, summarization, text cleaning, etc. GLiNER is built on top of encoder architecture such as BERT. In our case, we used DeBERTA v3 large \citep{he2023debertav3}, which improved the original DeBERTA model \citep{he2020deberta} with more efficient pertaining, replacing mask language modelling (MLM) with replaced token detection (RTD).

The main advantage of the GLiNER model is that it represents labels and text through a single forward path in the same encoder model. This enables the exchange of information between labels and text in both directions through the attention mechanism of a transformer. After passing through transformers, we extract the embedding of each label and token itself. Tokens embeddings additionally passed through the bidirectional LSTM model \citep{hochreiter1997long}. From our experiments, such an approach accelerates a model's training, so it is helpful in low data regimes; moreover, it limits the influence of negative tokenization and positional encoding artefacts. 

After getting representations of tokens and labels, we pass them through a scoring module that predicts the location of the token in an entity (beginning, inside, end) and its class. 

First of all, each token representation and label representation is projected into a higher-dimensional space.
Let $\mathbf{T} \in \mathbb{R}^{B \times L \times H}$ be the token representation matrix, where $B$ is the batch size, $L$ is the sequence length, and $H$ is the hidden size.
Let $\mathbf{L} \in \mathbb{R}^{B \times C \times H}$ be the label representation matrix, where $C$ is the number of classes.
The projections are defined as:
\[
\mathbf{T}' = \text{Linear}_T(\mathbf{T}) \in \mathbb{R}^{B \times L \times 2H}
\]
\[
\mathbf{L}' = \text{Linear}_L(\mathbf{L}) \in \mathbb{R}^{B \times C \times 2H}
\] 
After proper reshaping, we concatenate tokens and label representations with element-wise multiplication of tokens.

We expand and permute dimensions to align for concatenation in the following way:
\[
\mathbf{T}'' \rightarrow \mathbf{T}''' \in \mathbb{R}^{2 \times B \times L \times C \times H}
\]
\[
\mathbf{L}'' \rightarrow \mathbf{L}''' \in \mathbb{R}^{2 \times B \times L \times C \times H}
\]
We concatenate the representations along the last dimension:
\[
\mathbf{C} = \text{cat}(\mathbf{T}'''_0, \mathbf{L}'''_0, \mathbf{T}'''_1 \odot \mathbf{L}'''_1) \in \mathbb{R}^{B \times L \times C \times 3H}
\]

After that, we pass the combined representations through an MLP to produce the final scores for each class (start, end, score):
\[
\mathbf{S} = \text{MLP}(\mathbf{C}) \in \mathbb{R}^{B \times L \times C \times 3}
\]

To select the final set of output spans, we apply the same greedy decoding as used in the original GLiNER implementation which was deeply investigated in this work\citep{Zaratiana2022NamedER}. We use the average of the inside scores as the span score:
\[
\phi(i, j, c) = \frac{1}{j - i + 1} \sum_{k=i}^{j} \phi_i(k, c)
\]
Where $\phi(i, j, t)$ represents the span score for a span starting at position $i$ and ending at position $j$ for the token class $c$. The function $\phi_i(k, c)$ is the inside score for the token at position $k$ within the span, indicating the likelihood of the token being part of the entity class $c$. By averaging these scores over the span, we obtain a measure of how well the span fits the token class $c$.
\subsection{Data}
In this work, we generated a synthetic dataset using the Llama3 8B model processing English Wikipedia articles. Given a random article, the model was prompted to perform the following tasks:
\begin{itemize}[leftmargin=*, label={}]
    \item \textbf{Named Entity Recognition (NER)} - identification and categorization of entities such as names, organizations, dates, and other specific items in the text.
    \item \textbf{Open NER} - identification and categorization of entities based on provided specification by a user, for example, extraction of all products of a specific company.
    \item \textbf{Relation Extraction}: Detection and classification of relationships between entities within the text.
    \item \textbf{Summarization} - extraction of the most important sentences that summarize the input text, capturing the essential information.
    \item \textbf{Question-answering}: Finding an answer in the text given a question.
    \item \textbf{Open Information Extraction} - extraction of pieces of text given an open prompt from a user, for example, product description extraction.
\end{itemize}

Additionally, for tasks like open NER, summarization, question-answering, and open information extraction among an output, the model was asked to generate a prompt for the task given the context of a text. 

For each task, the LLM's output was processed to align it with the GLiNER format. Each example contains tokenized text and a list of spans, where we put the start and end token indexes together with labels. In the case of relations extraction, labels indicate the concatenation of source entity and relation, while for open information extraction, the label "match" was chosen. For question-answering, we use the label "answer". 

Another more high-quality dataset was used for post-fine-tuning. It combines half of the synthetic dataset and half of the manually curated NER datasets. 

\subsection{Models training}
The model training consisted of two distinct stages. Initially, we fine-tuned the model on a large synthetic dataset. Following this, we performed additional fine-tuning on a high-quality dataset, comprising 50\% named entity recognition (NER) examples and 50\% examples from other tasks.

In the first fine-tuning stage, the model was trained for 120,000 steps with a batch size of 8. The learning rate for the encoder was set to \( 1 \times 10^{-5} \), while the rest of the model had a learning rate of \( 5 \times 10^{-5} \). Both the encoder and the rest of the model utilized a weight decay of 0.01. We employed the Adam optimizer with the default parameters from PyTorch. The learning rate scheduler used was the cosine annealing scheduler. The training of the model employed a binary cross-entropy loss function, with a weighting scheme that assigned a weight of 0.75 to positive examples and 0.25 to negative examples. Each example during the training of the GLiNER model was limited to 30 labels, with a maximum sequence length of 768 words.

In the subsequent training phase, the model was further fine-tuned with a learning rate of \( 5 \times 10^{-6} \) for the encoder and \( 7 \times 10^{-6} \) for the rest of the model. A linear scheduler was applied in this phase. The training was conducted for an additional 1,000 steps, maintaining the same batch size as before.

Additionally, we have investigated the self-training capabilities of our model and other GLiNER models. We automatically pre-annotated a dataset from the domain NER benchmark and fine-tuned the model on it using the same learning rates as in the second training phase. To further enhance the model's performance, we introduced a label smoothing parameter into the loss function, which had a positive impact on the final performance of the model.

Label smoothing is a regularization technique used to improve the generalization of classification models by modifying the target probability distribution \citep{szegedy2015rethinking}. Instead of assigning a probability of 1 to the correct class and 0 to all others, label smoothing assigns a slightly lower probability to the correct class and distributes the remaining probability across the other classes. This technique helps to prevent the model from becoming overconfident in its predictions and can reduce overfitting.

For binary classification, we smoothed our targets in the following way:

\begin{equation}
\text{targets} \leftarrow \text{targets} \times (1 - \alpha) + 0.5 \times \alpha
\end{equation}

where $\alpha$ is the label smoothing parameter.

\subsection{Evaluation}
\subsubsection{Named Entity Recognition}
The model was compared with other GLiNER-type models, including both span-based and token-based models, such as NuMind-Zero \citep{bogdanov2024nuner}. All models were evaluated on cross-domain NER benchmarks in a zero-shot setting. We documented Micro-F1 scores along with precision and recall across all datasets. The datasets encompass diverse domains including AI, Literature, Politics, Science, Movies, and Restaurants. A threshold of 0.5 was applied to filter predicted entities, and all evaluations were conducted at the span level. Results are presented in Table \ref{tab:NER-Performance} and Results and Discussion section.

\subsubsection{Question-Answering}
The question-answering task was evaluated on the SQuAD2.0 dataset test subset \citep{rajpurkar-etal-2016-squad, rajpurkar-etal-2018-know} with the official evaluation script for SQuAD version 2.0, which calculates the exact match and F1 scores between predicted and actual values. We compared GLiNER multi-task with the models, capable of solving such a task, presented in Table \ref{tab:Validated-Models}. The threshold for filtering predicted entities by GLiNER multi-task, UTC collection was set to 0.5, and the simple aggregation strategy was used for input: "\{question\}\textbackslash{}n\{context\}". Flan-T5 models were tested with the default settings and the following input format: "Question: \{question\}\textbackslash{}nRules: As an answer, use only text from the context. Generate absolutely no text sequences not provided in the context, including punctuation marks that are not presented in the context.\textbackslash{}nContext: \{context\}". Meta-Llama-3-8B-Instruct was tested with the temperature 0.1 and the same aggregation strategy as Flan-T5 models.

\renewcommand{\arraystretch}{1.5}
\begin{table}[htbp]
\centering
\caption{Validated Models}
\label{tab:Validated-Models}
\begin{tabularx}{\textwidth}{l >{\raggedright\arraybackslash}p{2.5cm} >{\raggedright\arraybackslash}p{4.5cm} >{\raggedright\arraybackslash}X}
\hline
\textbf{Model} & \textbf{Size (params)} & \textbf{Architecture type} & \textbf{Reference}\\
\hline
gliner-multitask-large-v0.5 & 440M & GLiNER Encoder Transformer & -\\
UTC-DeBERTa-large-v2 & 434M & Encoder Transformer & \citep{knowledgator2024utc}\\
UTC-DeBERTa-base-v2 & 184M & Encoder Transformer & \citep{knowledgator2024utc}\\
UTC-DeBERTa-small-v2 & 141M & Encoder Transformer & \citep{knowledgator2024utc}\\
Flan-T5-small & 77M & Encoder-Decoder Transformer & \citep{https://doi.org/10.48550/arxiv.2210.11416}\\
Flan-T5-base & 248M & Encoder-Decoder Transformer & \citep{https://doi.org/10.48550/arxiv.2210.11416}\\
Flan-T5-large & 783M & Encoder-Decoder Transformer & \citep{https://doi.org/10.48550/arxiv.2210.11416}\\
Meta-Llama-3-8B-Instruct & 8.03B & Decoder Transformer & \citep{llama3modelcard}\\
\hline
\end{tabularx}
\end{table}
\renewcommand{\arraystretch}{1}

\subsubsection{Summarization}
Models in Table \ref{tab:Validated-Models} were tested for summarization task on the first 1k examples from the CNN Dailymail dataset \citep{DBLP:conf/nips/HermannKGEKSB15, see-etal-2017-get}. For GLiNER multi-task and UTC-collection we used a 0.1 threshold to filter predicted entities and the following input format: "Summarize the given text, highlighting the most important information:\textbackslash{}n\{context\}". For all generative models we used prompt: "Prompt: Summarize the given text, highlighting the most important. \textbackslash{}nText: \{text\}" and temperature 0.1 for Meta-Llama-3-8B-Instruct. Predicted strings were compared to human-created summaries from the dataset with ROUGE-1, ROUGE-2, and ROUGE-L scores. The standard deviation was calculated to understand data distribution and variability, which could indirectly point out the model stability on different data.

\subsubsection{Relation Extraction}
We also tested models in Table \ref{tab:Validated-Models} for the relation extraction task on the FewRel dataset "val\_wiki" subset \citep{han-etal-2018-fewrel, gao-etal-2019-fewrel}. GLiNER multi-task was tested with the input format: "Identify the relation in the given text, highlighting the relevant entity: \{text\}" and labels format: "\{head\} <> \{relation\}". To predict tails or objects with UTC collection we employed the following input format: "Identify target entity given the following relation: '\{relation\}' and the following source entity: '\{head\}' \textbackslash{}nText: \{text\}". For Flan-T5 collection we used: "Subject and relation:\{head\} <> \{relation\} \textbackslash{}n Context: \{text\}" as an input format. Meta-Llama-3-8B-Instruct was evaluated with the prompt: "As a target entity in answer, use only text from the text. Generate absolutely no sequences not provided in the text, including punctuation marks that are not presented in the text. Identify target entity given the following relation: '\{relation\}' and the following source entity: '\{head\}' \textbackslash{}nText: \{text\}". Predicted target entities were compared with the actual values with the exact match and F1 scores.

\subsubsection{Self-learning}
We tested several GLiNER models on cross-domain NER datasets after a single iteration of the self-learning procedure and observed improvements for most models across the majority of datasets. Multiple experiments were conducted, varying hyperparameters such as the datasets used for self-learning, learning rates, and loss function parameters like alpha and gamma. Additionally, we adjusted the label smoothing parameter, which significantly impacted model performance. The best results, along with the optimal hyperparameters, are documented and presented in the Results and Discussion section.

\section{Results and Discussion}\label{seq:results_and_discussion}
\subsection{Named Entity Recognition Results}
\begin{figure}[htbp]
    \centering
    \includegraphics[width=\textwidth]{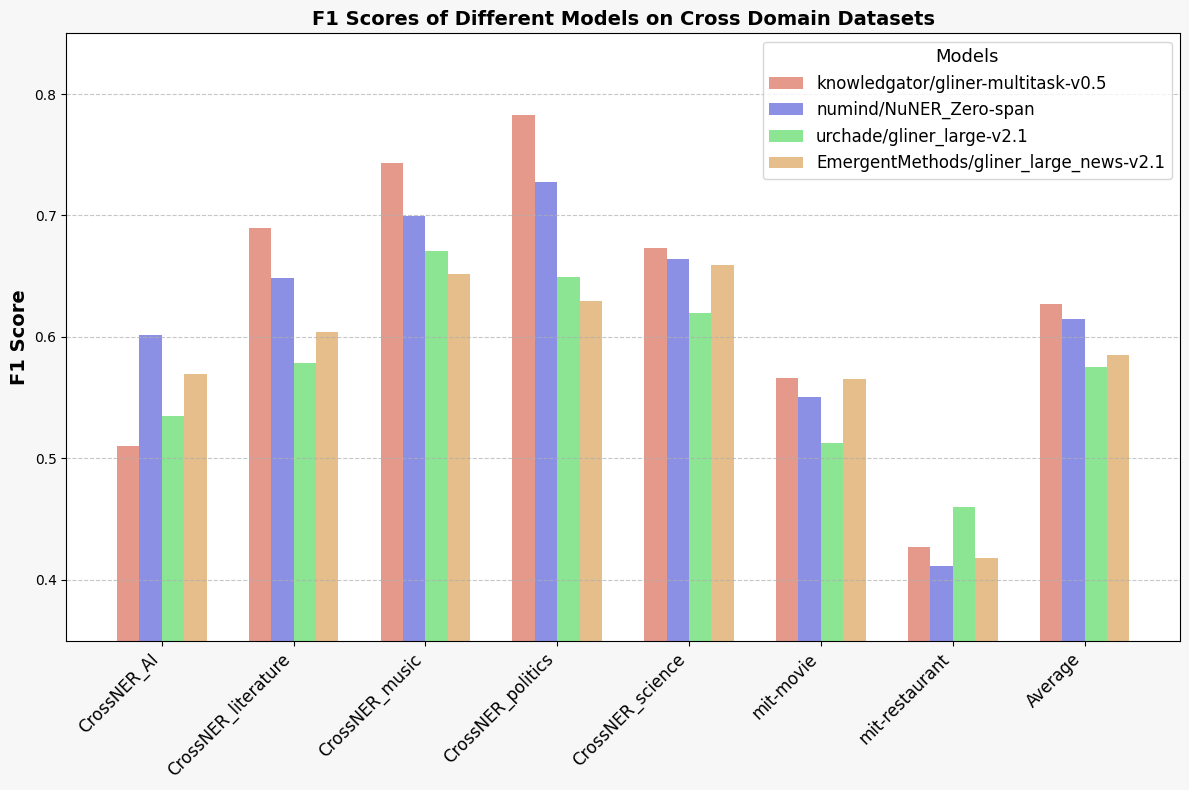}
    \caption{F1 scores for Named Entity Recognition tasks across different models and datasets.}
    \label{fig:ner-f1}
\end{figure}

In NER benchmarking, our model consistently outperformed other GLiNER models, particularly excelling in topics such as politics and literature \ref{fig:ner-f1}. However, it initially struggled with AI-related topics. Through experimentation with self-training, we managed to significantly improve its performance on the AI dataset. Notably, while all other models are NER-specific, our model was trained on a diverse range of tasks. This suggests that even relatively compact encoder models for our days like DeBERTa-large, which serves as the backbone for all tested models, can achieve competitive performance across various tasks. This underscores the potential benefits of transfer learning for specific tasks like named entity recognition.
\renewcommand{\arraystretch}{1}
\begin{table}[htbp]
\centering
\caption{Models Performance in NER task}
\label{tab:NER-Performance}
\begin{tabularx}{\textwidth}{p{5cm} p{3cm} X X X} % Adjust the width of p{3.5cm} as needed.
\hline
\textbf{Model} & \textbf{Dataset} & \textbf{Precision} & \textbf{Recall} & \textbf{F1 Score }\\
\hline
\texttt{gliner-multitask-v0.5} & CrossNER\_AI & 51.00\% & 51.11\% & 0.5105 \\
& CrossNER\_literature & 72.65\% & 65.62\% & 0.6896 \\
& CrossNER\_music & 74.91\% & 73.70\% & 0.7430 \\
& CrossNER\_politics & 78.84\% & 77.71\% & 0.7827 \\
& CrossNER\_science & 69.20\% & 65.48\% & 0.6729 \\
& mit-movie & 61.29\% & 52.59\% & 0.5660 \\
& mit-restaurant & 50.65\% & 38.13\% & 0.4351 \\
& \textbf{Average} & & & \textbf{0.6276} \\
\hline
\texttt{NuNER\_Zero-span} & CrossNER\_AI & 63.82\% & 56.82\% & 0.6012 \\
& CrossNER\_literature & 73.53\% & 58.06\% & 0.6489 \\
& CrossNER\_music & 72.69\% & 67.40\% & 0.6995 \\
& CrossNER\_politics & 77.28\% & 68.69\% & 0.7273 \\
& CrossNER\_science & 70.08\% & 63.12\% & 0.6642 \\
& mit-movie & 63.00\% & 48.88\% & 0.5505 \\
& mit-restaurant & 54.81\% & 37.62\% & 0.4462 \\
& \textbf{Average} & & & \textbf{0.6196} \\
\hline
\texttt{gliner-large-news-v2.1} & CrossNER\_AI & 59.60\% & 54.55\% & 0.5696 \\
& CrossNER\_literature & 65.41\% & 56.16\% & 0.6044 \\
& CrossNER\_music & 67.47\% & 63.08\% & 0.6520 \\
& CrossNER\_politics & 66.05\% & 60.07\% & 0.6292 \\
& CrossNER\_science & 68.44\% & 63.57\% & 0.6592 \\
& mit-movie & 65.85\% & 49.59\% & 0.5657 \\
& mit-restaurant & 54.71\% & 35.94\% & 0.4338 \\
& \textbf{Average} & & & \textbf{0.5876} \\
\hline
\texttt{gliner\_large-v2.1} & CrossNER\_AI & 54.98\% & 52.00\% & 0.5345 \\
& CrossNER\_literature & 59.33\% & 56.47\% & 0.5787 \\
& CrossNER\_music & 67.39\% & 66.77\% & 0.6708 \\
& CrossNER\_politics & 66.07\% & 63.76\% & 0.6490 \\
& CrossNER\_science & 61.45\% & 62.56\% & 0.6200 \\
& mit-movie & 55.94\% & 47.36\% & 0.5129 \\
& mit-restaurant & 53.34\% & 40.83\% & 0.4625 \\
& \textbf{Average} & & & \textbf{0.5754} \\
\hline
\end{tabularx}
\end{table}
\renewcommand{\arraystretch}{1}

\subsection{Question-Answering Results}
GLiNER multi-task achieved top results in the question-answering task (Table \ref{tab:Question-Answering-SQuAD2.0}) with the exact match score of 87.72, but UTC-DeBERTa-large-v2 achieved a higher F1 score, 92.53 versus 91.99. Generative models reached comparatively low scores, highlighting the instability of models on input data for such a task, especially with the growth of model size. 
\renewcommand{\arraystretch}{1.5}
\begin{table}[htbp]
\centering
\caption{Question-Answering SQuAD2.0}
\label{tab:Question-Answering-SQuAD2.0}
\begin{tabularx}{\textwidth}{l X X X} % Use tabularx with \textwidth and adjust column types accordingly.
\hline % Use \hline for horizontal lines in basic tabular or tabularx environments.
\textbf{Model} & \textbf{Size (params)} & \textbf{Exect match} & \textbf{F1 score}\\
\hline % Replace \toprule and \midrule from booktabs with \hline 
gliner-multitask-large-v0.5&440M&87.72&91.99 \\
UTC-DeBERTa-large-v2&434M &86.06&92.53 \\
UTC-DeBERTa-base-v2 &184M&83.95&90.18 \\
UTC-DeBERTa-small-v2 &141M&80.92&86.89 \\
Flan-T5-small&77M&79.68&85.86 \\
Flan-T5-base&248M&83.39&90.33 \\
Flan-T5-large&783M&81.05&89.39 \\
Meta-Llama-3-8B-Instruct&8.03B&68.94&80.51 \\
\hline % Replace \bottomrule from booktabs with \hline when using vertical lines.
\end{tabularx}
\end{table}
\renewcommand{\arraystretch}{1}

\subsection{Summarization Results}
GLiNER multi-task outperformed all models (Table \ref{tab:Summarization-CNN-DailyMail}) from the list with 0.2484±0.1142 ROUGE-1, 0.0881±0.0892 ROUGE-2, and 0.2279±0.1117 ROUGE-L scores.
\renewcommand{\arraystretch}{1.5}
\begin{table}[htbp]
\centering
\caption{Summarization CNN DailyMail}
\label{tab:Summarization-CNN-DailyMail}
\begin{tabularx}{\textwidth}{l X X X} % Use tabularx with \textwidth and adjust column types accordingly.
\hline % Use \hline for horizontal lines in basic tabular or tabularx environments.
\textbf{Model} & \textbf{ROUGE-1 and Std} & \textbf{ROUGE-2 and Std} & \textbf{ROUGE-L and Std}\\
\hline % Replace \toprule and \midrule from booktabs with \hline 
gliner-multitask-large-v0.5 &0.2484±0.1142&0.0881±0.0892&0.2279±0.1117\\
UTC-DeBERTa-large-v2 &0.2409±0.1145&0.0785±0.0870&0.2164±0.1091 \\
UTC-DeBERTa-base-v2 &0.2143±0.0983&0.0603±0.0713&0.1955±0.0928\\
UTC-DeBERTa-small-v2&0.1813±0.1258&0.0510±0.0800&0.1618±0.1172 \\
Flan-T5-small&0.1935±0.1114&0.0565±0.0784&0.1767±0.1041 \\
Flan-T5-base&0.2166±0.1232&0.0693±0.0899&0.1983±0.1194 \\
Flan-T5-large&0.2055±0.1253&0.0676±0.0969&0.1876±0.1211 \\
Meta-Llama-3-8B-Instruct& 0.2160±0.0708&0.0644±0.0465&0.1991±0.0663 \\
\hline % Replace \bottomrule from booktabs with \hline when using vertical lines.
\end{tabularx}
\end{table}
\renewcommand{\arraystretch}{1}

\subsection{Relation Extraction Results}
In the Relation Extraction task, GLiNER multi-task model showcased impressive results with 82.5 exact matches and 87.36 F1 scores. Because negatives were sampled from a batch with examples that belong to other tasks, better sampling strategies for hard negatives should achieve even better results. In our future works, we will explore this direction as well as the end-to-end relation extraction capabilities of GLiNER models.
\renewcommand{\arraystretch}{1.5}
\begin{table}[htbp]
\centering
\caption{Relation Extraction FewRel}
\label{tab:Relation-Extraction-FewRel}
\begin{tabularx}{\textwidth}{l X X X} % Use tabularx with \textwidth and adjust column types accordingly.
\hline % Use \hline for horizontal lines in basic tabular or tabularx environments.
\textbf{Model} & \textbf{Size (params)} & \textbf{Exect match} & \textbf{F1 score}\\
\hline % Replace \toprule and \midrule from booktabs with \hline 
gliner-multitask-large-v0.5&440M&82.5&87.36 \\
UTC-DeBERTa-large-v2&434M &71.91&80.95 \\
UTC-DeBERTa-base-v2 &184M&64.06&72.32 \\
UTC-DeBERTa-small-v2 &141M&48.39&55.95 \\
Flan-T5-small&77M&22.25&29.45 \\
Flan-T5-base&248M&53.75&60.21 \\
Flan-T5-large&783M&55.30&61.20 \\
Meta-Llama-3-8B-Instruct&8.03B&38.28&44.28 \\
\hline % Replace \bottomrule from booktabs with \hline when using vertical lines.
\end{tabularx}
\end{table}
\renewcommand{\arraystretch}{1}

\subsection{Self-training Results}
 Overall, the average improvement in F1 score reached up to 2\% on the cross-domain NER benchmark(Table \ref{tab:Self-Learning-Performance}). More interestingly the major improvements were observed for datasets where the initial model demonstrated relatively poor performance. For example, in the case of our multi-task GLiNER model, the initial F1 score on the CrossNER AI dataset was 0.5105, while after the self-learning procedure, it reached 0.6325. For the cases when the model demonstrated already relatively good performance (more than 0.6 F1 score), we observed no difference or slight decrease in performance.
\renewcommand{\arraystretch}{1.5}
\begin{table}[htbp]
\centering
\caption{Self-Learning Performance Comparison for Different GLiNER Models}
\label{tab:Self-Learning-Performance}
\begin{tabular}{l *{8}{c}} % Changed tabularx to tabular and X to c
\hline
\textbf{Model} & \rotatebox{90}{\textbf{Steps}} & \rotatebox{90}{\textbf{Alpha}} & \rotatebox{90}{\textbf{Gamma}} & \rotatebox{90}{\textbf{LR (encoder)}} & \rotatebox{90}{\textbf{LR (other)}} & \rotatebox{90}{\textbf{Label Smoothing }} & \rotatebox{90}{\textbf{Initial F1}} & \rotatebox{90}{\textbf{Final F1}} \\
\hline
gliner-multitask-large & 500 & 0.75 & 0 & $5.00 \times 10^{-6}$ & $7.00 \times 10^{-6}$ & 0.2 & 0.6276 & 0.6416 \\
urchade/gliner\_large-v2.1 & 1000 & 0.75 & 0 & $5.00 \times 10^{-6}$ & $5.00 \times 10^{-6}$ & 0.01 & 0.5754 & 0.59237 \\
numind/NuNER\_Zero-span & 100 & 0.75 & 0 & $5.00 \times 10^{-6}$ & $5.00 \times 10^{-6}$ & 0.01 & 0.6196 & 0.6295 \\
\hline
\end{tabular}
\end{table}
\renewcommand{\arraystretch}{1}

\subsection{Discussion}

In this study, we explored a GLiNER-based token classification architecture, utilizing the powerful DeBERTa v3 encoder model, across a spectrum of information extraction tasks. Our findings demonstrate the effectiveness of our approach in tasks such as Named Entity Recognition (NER), Question-Answering, Summarization, and Relation Extraction, showcasing its versatility and efficiency in handling structured output requirements.

Our investigation of synthetic data generation using large language models (LLMs) like Llama3 8B revealed its potential in creating diverse and high-quality datasets for NLP model training. This approach enabled us to train our encoder-based model on a wide range of tasks, significantly improving its generalization capabilities. Notably, our model surpassed its teacher model on several datasets. It indicated that the open generation of label data by LLM produces better results than labelling with a fixed set of classes and other constraints settled by a user. 

The architecture of our model also played a crucial role in its performance. Encoder-based models offer several advantages over decoders, including bi-directional attention mechanisms, output efficiency, and consistency. Additionally, the GLiNER architecture's ability to handle labels and text in a single forward pass sets it apart from similar approaches like UTC. The bidirectional nature of encoders facilitates bidirectional communication between labels and text, leading to better representations for both labels and tokens compared to bi-encoder architectures.

Furthermore, our exploration of self-training techniques with our model and other GLiNER models showcased performance improvements through iterative self-learning procedures. This highlights the potential of self-learning techniques in enhancing model performance, particularly in scenarios with limited labelled data, thus underscoring their significance in real-world applications.

Overall, our work demonstrates the effectiveness of leveraging various techniques, including synthetic data generation, multi-task learning, and self-learning, to enhance the performance of NLP models across a range of information extraction tasks. These findings contribute to the advancement of AI research and hold promise for impacting diverse real-world applications.

\section{Conclusion}
The GLiNER multi-task model demonstrated strong generalization across various information extraction tasks, including named entity recognition (NER), relation extraction, summarization, and question answering. While further research is needed to fully understand its ability to comprehend a diverse range of prompts, it has already shown significant potential for transferability to numerous critical applications. Overall, we have shown that even relatively small encoder models fine-tuned on large and diverse datasets can exhibit good prompt-tuning abilities and achieve good performance across information extraction tasks. Usage of LLMs can be ineffective for such tasks due to their expensive inference and tendency to hallucinate and fail to provide structural outputs. Interestingly, the GLiNER multi-task model outperformed its teacher model, Meta-Llama-3-8B-Instruct, highlighting its superior performance. This underscores the efficacy of synthetic data generation and the robustness of encoder-like models. The architecture of GLiNER provides enhanced scalability, greater control over outputs, and improved interoperability. The success of this approach opens up several avenues for future research. We plan to continue exploring this direction, with a particular focus on investigating the scaling properties of such multi-task encoder models. Additionally, we aim to create more diverse and larger high-quality datasets to further improve the model's performance and generalization capabilities.
\newpage
\section{Availability}
The model is available for use with the Python library GLiNER at 
 \url{https://github.com/urchade/GLiNER}, and also through the framework UTCA for building pipelines at \url{https://github.com/Knowledgator/utca}. The model can be downloaded from the Hugging Face repository at 
 \url{https://huggingface.co/knowledgator/gliner-multitask-large-v0.5}. For quick testing of the model, we recommend using the demo application GLiNER HandyLab at \url{https://huggingface.co/spaces/knowledgator/GLiNER_HandyLab}, which contains templates for all tasks tested in this work.
\section{Acknowledgments}
We express our gratitude to Urchade Zaratiana, the developer of GLiNER, for his invaluable support in the release of the GLiNER multi-task model.
%Bibliography
\bibliographystyle{plainnat}  
\bibliography{main}  
\end{document}